\documentclass[pdflatex,sn-mathphys-num]{sn-jnl}%

\usepackage{graphicx}%
\usepackage{multirow}%
\usepackage{amsmath,amssymb,amsfonts}%
\usepackage{amsthm}%
\usepackage{mathrsfs}%
\usepackage[title]{appendix}%
\usepackage{xcolor}%
\usepackage{textcomp}%
\usepackage{manyfoot}%
\usepackage{booktabs}%
\usepackage{algorithm}%
\usepackage{algorithmicx}%
\usepackage{algpseudocode}%
\usepackage{listings}%
\usepackage{booktabs} %
\usepackage{placeins}

\theoremstyle{thmstyleone}%
\theoremstyle{thmstyletwo}%

\theoremstyle{thmstylethree}%

\begin{document}

\title[Clinical DVH Metric Loss]{Clinical DVH metrics as a loss function for 3D dose prediction in head and neck radiotherapy}




\author*[1]{\fnm{Ruochen} \sur{Gao}}\email{r.gao@lumc.nl}

\author[1,2]{\fnm{Marius} \sur{Staring}}

\author[2]{\fnm{Frank} \sur{Dankers}}

\affil[1]{\orgdiv{Division of Image Processing, Department of Radiology}, \orgname{Leiden University Medical Center}, \orgaddress{ \city{Leiden}, \country{the Netherlands}}}

\affil[2]{\orgdiv{Department of Radiation Oncology}, \orgname{Leiden University Medical Center}, \orgaddress{ \city{Leiden}, \country{the Netherlands}}}

\abstract{\textbf{Purpose} 
Deep-learning-based three-dimensional (3D) dose prediction has become an important component of automated radiotherapy workflows. However, most existing models are trained using voxel-wise regression losses, which are poorly aligned with clinical plan evaluation criteria that rely on dose-volume histogram (DVH)-derived metrics. This study aims to develop a clinically guided loss formulation that directly optimizes clinically used DVH metrics while remaining computationally efficient for head and neck (H\&N) dose prediction.

\textbf{Methods} 
We propose a clinical DVH metric loss (CDM loss) that jointly incorporates differentiable formulations of common \textit{D--metrics} and differentiable surrogates of \textit{V--metrics}. The loss is driven by a JSON-based clinical plan evaluation template to ensure alignment with clinic planning criteria. In addition, we introduce a lossless bit-mask encoding scheme to efficiently represent a large number of overlapping ROIs, substantially improving training efficiency. The method was evaluated on a retrospective cohort of 174 H\&N patients treated with VMAT, using a temporal split for training ($n = 137$) and testing ($n = 37$).

\textbf{Results} 
Compared with MAE- and DVH-based losses, the proposed CDM loss substantially improved clinical target coverage, and satisfied all predefined clinical constraints. Using a standard 3D U-Net, the PTV Score was reduced from 1.544 (MAE) to 0.491 (MAE+CDM), while consistently satisfying all PTV clinical constraints. OAR sparing remained comparable or improved. Bit-mask ROI encoding reduced average training epoch time from 241~s to 43~s (83.2\% reduction) and lowered peak GPU memory usage. Notably, with the CDM loss, a standard 3D U-Net achieved performance comparable to or better than more complex state-of-the-art architectures.

\textbf{Conclusion} 
Directly optimizing clinically used DVH metrics enables 3D dose predictions that are better aligned with clinical treatment planning criteria than conventional only voxel-wise or DVH-curve supervision. The proposed CDM loss, combined with efficient ROI bit-mask encoding, provides a practical and scalable framework for H\&N dose prediction.}




\maketitle

\section{Introduction}\label{intro}

Radiotherapy (RT) is widely used in cancer treatment to deliver sufficient radiation to eradicate tumor cells while protecting surrounding healthy tissues. Conventional treatment planning directly optimizes beam parameters within the treatment planning system (TPS), often requiring multiple rounds of manual refinements and substantial planner expertise. This process is highly time-consuming, especially for anatomically complex sites. To improve efficiency and consistency, modern workflows increasingly introduce dose prediction as an intermediate step, providing an anatomy-informed estimate of a clinically realistic three-dimensional (3D) dose distribution. This predicted dose can then be used to guide subsequent TPS optimization through a dose mimicking process, in which a deliverable treatment plan is optimized to closely reproduce the predicted 3D dose distribution~\cite{bakx2021development}.

Among all treatment sites, head and neck (H\&N) radiotherapy is particularly challenging due to its intricate anatomy, a large number of critical organs-at-risk (OARs), and the need for steep dose gradients between planning target volumes (PTVs) and nearby healthy organs~\cite{buciuman2022adaptive}. These complexities make high-quality optimization particularly demanding and contribute to considerable variability in plan quality across clinicians and institutions~\cite{berry2016interobserver}. Therefore, developing accurate and automated 3D dose prediction models tailored for H\&N treatment is essential to reduce planning workload and promote standardization in clinical practice.

\begin{figure}[tb]
    \centering
    \includegraphics[scale=0.40]{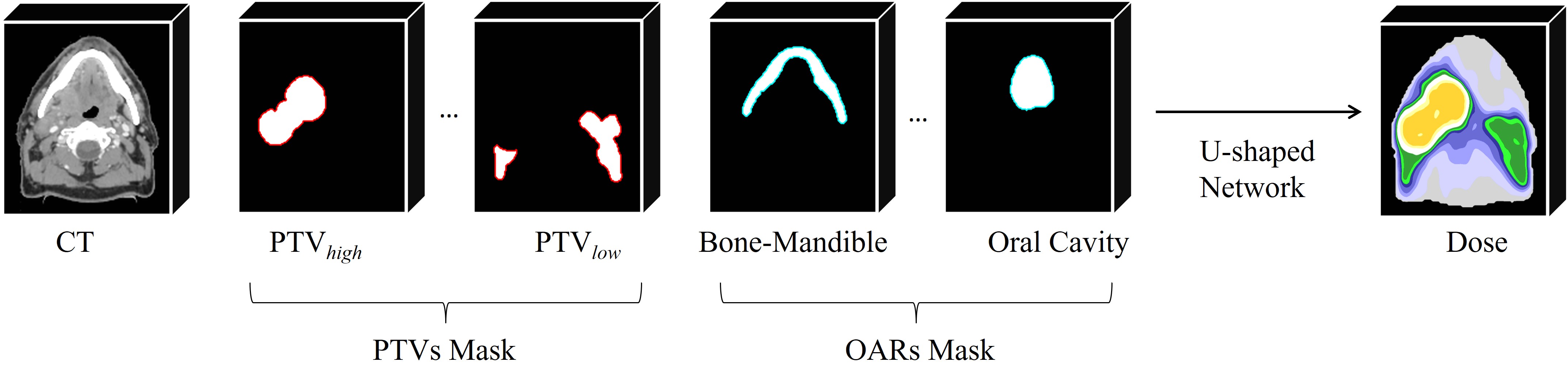}
    \caption{Workflow of a typical deep-learning-based H\&N dose prediction method. The planning CT and masks of the PTVs and OARs are provided as inputs to a U-shaped neural network to generate a 3D voxel-wise dose distribution. In H\&N cases, the PTVs are typically assigned two or three dose levels, while the number of OARs is relatively large compared to other treatment sites.}
\label{fig:Intro}
\end{figure}

Early dose prediction approaches relied on atlas-based and knowledge-based planning, which generated 3D dose estimates from prior clinical plans and handcrafted anatomical features~\cite{mcintosh2015contextual}. Although these methods improve planning consistency, they are limited in capturing the complex nonlinear relationship between anatomy and dose.
Recently, deep learning (DL) methods have shown strong potential for learning complex mappings between patient anatomy and clinical dose patterns. Most DL-based methods use 3D convolutional or transformer-based U-shaped architectures~\cite{liu2021cascade,lin2024lenas,gheshlaghi2024cascade,gao2025factors}, taking CT and region–of–interest (ROI) masks as input to generate 3D voxel-wise dose predictions, as shown in Fig.~\ref{fig:Intro}. These approaches generally achieve high voxel-wise accuracy because they are trained with regression losses such as mean absolute error (MAE) or mean squared error (MSE), which encourage global similarity to clinical dose distributions.
However, voxel-wise accuracy is not the primary criterion used in clinical plan evaluation. Instead, clinicians focus on achieving adequate planning target volume (PTV) coverage and sufficient organ-at-risk (OAR) sparing, which are typically assessed using dose–volume histogram (DVH)-derived metrics. Because standard regression losses do not directly optimize these clinical objectives, DL-based models may achieve high numerical accuracy, yet still perform sub-optimally from a clinical standpoint.

To bridge this gap, several studies have incorporated differentiable DVH losses, which allow comparison of predicted and ground-truth DVH curves during training~\cite{nguyen2020incorporating, jhanwar2022domain, wang2022deep, teng2024beam}. In clinical practice, however, plan evaluation is typically based on a collection of dose metrics rather than full DVH curves. For H\&N RT, these metrics commonly include both \textit{D--metrics} 
(e.g., $D_{x\%}$, the minimum dose received by $x$\% of an ROI) and \textit{V--metrics}
(e.g., $V_{x\%}$, the volume of an ROI receiving at least $x\%$ of the prescription dose), as summarized in Table~\ref{tab:DV_definitions}. While \textit{D--metrics} are differentiable and can be incorporated into training~\cite{wang2022deep, teng2024beam}, \textit{V--metrics} contain threshold operations that are inherently non-differentiable, posing challenges for the gradient-based training framework. Therefore, effective methods for directly optimizing \textit{V--metrics} remain limited, and a comprehensive framework that jointly supports both \textit{D–} and \textit{V--metrics} is still needed.

\begin{table}[tb]
\caption{Definitions of common \textit{D--} and \textit{V--metrics} in H\&N RT.}
\label{tab:DV_definitions}
\centering
\renewcommand{\arraystretch}{1.1}
\begin{tabular}{ll}
\toprule
\textbf{Parameter} & \textbf{Definition} \\
\midrule
$D_{x\%}$ & Minimum dose received by $x$\% of the ROI volume. \\
$D_{x~\text{cc}}$ & Minimum dose received by the hottest $x$~cc of the ROI. \\ 
$D_{\text{max}}$ & Maximum dose within the ROI. \\ 
$D_{\text{min}}$ & Minimum dose within the ROI. \\
$D_{\text{mean}}$ & Mean dose within the ROI. \\
$V_{x\%}$ & Volume percentage of an ROI receiving at least $x$\% of the prescription dose. \\
$V_{x~\text{Gy}}$ & Volume percentage of an ROI receiving at least $x$~Gy. \\
\bottomrule
\end{tabular}
\end{table}

Another practical challenge arises from the large number of OARs in H\&N anatomy. Unlike DL-based medical imaging segmentation models, which often train on small 3D patches, dose–prediction networks typically require the full 3D volume as input to guarantee spatial dose continuity. Current methods generally implement these ROIs as one input channel per ROI. As a result, when many OARs are included, the number of input channels grows rapidly, leading to substantial preprocessing overhead, higher CPU–-to–-GPU data transfer cost, increased GPU memory usage, and longer training time.
 Although merging all ROIs into a single indexed mask could reduce input dimensionality, this approach faces two major issues: (1) many OARs overlap spatially (e.g., the brain and brainstem overlap), making direct merging impossible; and (2) the numerical labels introduce unintended ordinal bias (e.g., a label “9” may be treated differently from a label “1” purely because it is larger), even though ROI labels should carry no numeric significance. This unintended numerical meaning can mislead the network during training and negatively affect learning~\cite{hancock2020survey}. As a result, prior studies often limit the input to a subset of OARs~\cite{babier2021openkbp,lin2024lenas,gheshlaghi2024cascade,gao2025factors, wang2022deep, teng2024beam}, which is restrictive because clinical evaluation typically involves a much broader set of OARs. This challenge highlights the need for models that can efficiently incorporate all clinically relevant ROIs while maintaining computational efficiency.

To address these limitations, we propose a clinical DVH metric loss (CDM loss), together with an efficient bit–mask ROI encoding strategy for 3D H\&N dose prediction. Our framework enables direct optimization of the full set of clinically defined dose evaluation metrics while maintaining computational efficiency. Our main contributions are summarized as follows:

\begin{itemize}

\item We introduce a clinical DVH metric loss (CDM loss) that combines differentiable formulations for \textit{D--metrics} with differentiable approximations for \textit{V--metrics}. By directly optimizing the metrics used in clinical plan evaluation, we show that even a standard 3D U-Net can achieve clinically satisfactory dose predictions.

 \item We develop an efficient lossless bit–mask encoding for ROI representation that reduces CPU preprocessing, CPU–-to–-GPU data transfer, and GPU memory usage through lightweight, on-demand decoding. This strategy substantially accelerates training, while supporting the full set of OARs required for clinical evaluation.

\item We organize the clinical evaluation dose metrics used by CDM loss function through a simple JSON-based template, as summarized in Table~\ref{tab:clinical plan evaluation template}. This template specifies the ROIs and their associated dose metrics in a clear, structured, and easily updated format, ensuring that the training objectives remain aligned with clinical evaluation criteria used in a given institute.

\end{itemize}

\begin{table}[tb]
\caption{The H\&N RT clinical plan evaluation template used in our institution. 
The PTVs are denoted by their prescribed dose levels (e.g., $\text{PTV}_{54.25}$ for 54.25 Gy and $\text{PTV}_{70}$ for 70 Gy). 
ROIs with “L/R” denote paired organs and are evaluated independently for the left and right sides. Each ROI is evaluated using one or more dose metrics.
“Aim” represents the desired planning goal, while “Constraint” denotes a mandatory criterion.
The template is implemented via a JSON file.}
\label{tab:clinical plan evaluation template}
\centering
\renewcommand{\arraystretch}{1.1}
\setlength{\tabcolsep}{16pt}       
\begin{tabular}{llll} %

\toprule
\textbf{ROI} & \textbf{Metric} & \textbf{Aim} & \textbf{Constraint} \\
\midrule
$\text{PTV}_{54.25}$ & \(V_{95\%}\) &  & $\geq 98\%$ \\
$\text{PTV}_{54.25}$ & \(D_{\text{mean}}\) & $\leq 102\%$ &  \\
$\text{PTV}_{70}$ & \(V_{95\%}\) &  &$\geq 98\%$ \\
$\text{PTV}_{70}$ & \(D_{0.03\text{cc}}\) & $\leq 107\%$ & $\leq 110\%$ \\
$\text{PTV}_{70}$ & \(D_{\text{mean}}\) &  $\leq 102\%$& \\
SpinalCord & \(D_{0.03\text{cc}}\) &  & $\leq 50\,\text{Gy}$ \\
SpinalCord~+~3~mm & \(D_{0.03\text{cc}}\) &  & $\leq 52\,\text{Gy}$ \\
Brainstem\_Surf & \(D_{0.03\text{cc}}\) &  & $\leq 60\,\text{Gy}$ \\
Brainstem\_Core & \(D_{0.03\text{cc}}\) &  & $\leq 54\,\text{Gy}$ \\
Brain & \(D_{0.03\text{cc}}\) & $\leq 65\,\text{Gy}$ &  \\
Brain & \(D_{2\%}\) & $\leq 70\,\text{Gy}$ &  \\
Cochlea\_(L/R) & \(D_{\text{mean}}\) & $\leq 45\,\text{Gy}$ &  \\
Parotid\_(L/R) & \(D_{\text{mean}}\) & $\leq 28\,\text{Gy}$ &  \\
Glnd\_Submand\_(L/R) & \(D_{\text{mean}}\) & $\leq 35\,\text{Gy}$ &  \\
Oral\_Cavity & \(D_{\text{mean}}\) & $\leq 28\,\text{Gy}$ &  \\
Musc\_Constrict\_S & \(D_{\text{mean}}\) & $\leq 40\,\text{Gy}$ &  \\
Musc\_Constrict\_M & \(D_{\text{mean}}\) & $\leq 40\,\text{Gy}$ &  \\
Musc\_Constrict\_I & \(D_{\text{mean}}\) & $\leq 40\,\text{Gy}$ &  \\
Cricopharyngeus & \(D_{\text{mean}}\) & $\leq 40\,\text{Gy}$ &  \\
Larynx\_SG & \(D_{\text{mean}}\) & $\leq 40\,\text{Gy}$ &  \\
Glottic\_Area & \(D_{\text{mean}}\) & $\leq 40\,\text{Gy}$ &  \\
Bone\_Mandible & \(D_{2\%}\) & $\leq 70\,\text{Gy}$ &  \\
Bone\_Mandible-PTV & \(D_{2\%}\) & $\leq 50\,\text{Gy}$ &  \\
Eye\_(L/R) & \(D_{0.03\text{cc}}\) & $\leq 35\,\text{Gy}$ &  \\
Lens\_(L/R) & \(D_{0.03\text{cc}}\) & $\leq 6\,\text{Gy}$ &  \\
Pituitary & \(D_{\text{mean}}\) & $\leq 20\,\text{Gy}$ &  \\
OpticNrv\_(L/R) & \(D_{0.03\text{cc}}\) & $\leq 55\,\text{Gy}$ &  \\
\bottomrule
\end{tabular}
\end{table}

\section{Materials}\label{sec2}

\subsection{Dataset}
In this study, data were collected retrospectively from 174 H\&N cancer patients treated at Leiden University Medical Center from February 2021 to July 2025. The study was approved by the Medical Ethics Committee of Leiden, The Hague, and Delft (approval number G21.142, October 15, 2021). The cohort covered a range of tumor sites, including the oropharynx, hypopharynx, nasopharynx, larynx, and oral cavity. All patients were treated on Elekta Synergy linear accelerators (Elekta, Stockholm, Sweden) using 6 MV dual-arc, full-rotation volumetric modulated arc therapy (VMAT) and all RT plans were generated in RayStation$^\text{TM}$~(RaySearch Laboratories, Stockholm, Sweden).

To better reflect real-world clinical application, in which a model trained on historical data is used to make predictions for future patients, the dataset was divided chronologically. Patients treated between February 2021 and July 2024 ($n = 137$) were assigned to the training cohort, while those treated between August 2024 and July 2025 ($n = 37$) formed the test cohort, enabling temporal (out-of-time) validation on later cases.

The original CT images had an average resolution of $1.2 \times 1.2 \times 2\,\mathrm{mm}^3$, whereas the dose grid resolution was fixed at $2 \times 2 \times 2\,\mathrm{mm}^3$. To ensure uniform spatial resolution, all CT scans were resampled to $2 \times 2 \times 2\,\mathrm{mm}^3$. Cropping or padding were performed around the treatment isocenter, defined as the geometric center of the planned radiotherapy beams, to obtain a uniform volume size of $(160, 192, 256)$. The same preprocessing pipeline was applied to all ROIs.

For intensity normalization, CT values were clipped to the range $[-1024, 1024]$ and then normalized to $[0, 1]$ using min–max normalization. Dose values were normalized by dividing by 70 (Gy), the highest prescription dose in our cohort.

\subsection{Evaluation Metrics}

To quantitatively evaluate model performance,
three complementary evaluation metrics are defined: PTV Score, OAR Score, and Dose Score. The PTV Score and OAR Score are directly derived from the clinical plan evaluation template used in our institution (Table~\ref{tab:clinical plan evaluation template}), which specifies the clinical dose evaluation metrics for each ROI. These two scores are reported separately because they represent different clinical objectives and optimization priorities.

Let $\mathcal{P}$ denote the set of all PTVs, and $\mathcal{O}$ denote the set of all OARs. For each ROI $r \in \mathcal{P} \cup \mathcal{O}$, a corresponding set of clinical evaluation dose metrics $\mathcal{M}_r$ 
is defined according to Table~\ref{tab:clinical plan evaluation template}. For example, $\mathcal{M}_r = \{V_{95\%},~D_{\text{mean}},~D_{0.03\text{cc}}\}$ for $\text{PTV}_{70}$, and $\mathcal{M}_r = \{D_{0.03\text{cc}},~D_{2\%}\}$ for Brain. For each metric $m \in \mathcal{M}_r$, let $M^{\text{pred}}_{r,m}$ and $M^{\text{gt}}_{r,m}$ denote the predicted and ground truth (clinical) values, respectively.

The PTV Score is defined as:
\begin{equation}
\text{PTV Score} = 
\frac{1}{|\mathcal{P}|}
\sum_{p \in \mathcal{P}}
\frac{1}{|\mathcal{M}_p|}
\sum_{m \in \mathcal{M}_p}
\left|
M^{\text{pred}}_{p,m} - M^{\text{gt}}_{p,m}
\right|,
\end{equation}
where all PTV metrics are normalized to the prescribed dose 
and expressed in percentage (\%). Similarly, the OAR Score is defined as:
\begin{equation}
\text{OAR Score} = 
\frac{1}{|\mathcal{O}|}
\sum_{o \in \mathcal{O}}
\frac{1}{|\mathcal{M}_o|}
\sum_{m \in \mathcal{M}_o}
\left|
M^{\text{pred}}_{o,m} - M^{\text{gt}}_{o,m}
\right|,
\end{equation}
where all OAR metrics are measured in absolute dose (Gy). 

The Dose Score is also reported to quantify the overall voxel-wise difference between the predicted and ground truth (clinical) 3D dose distributions. Let $\Omega$ denote the set of all voxels within the dose grid, and $D^{\text{pred}}(v)$ and $D^{\text{gt}}(v)$ be the predicted and clinical dose at voxel $v \in \Omega$. The Dose Score is defined as:
\begin{equation}
\text{Dose Score} =
\frac{1}{|\Omega|}
\sum_{v \in \Omega}
\left|
D^{\text{pred}}(v) - D^{\text{gt}}(v)
\right|,
\end{equation}
reported in Gy.
Although this metric has limited direct clinical relevance, 
it is widely adopted in the literature and facilitates consistent comparison with prior studies.

\section{Method}

\subsection{CDM loss Formulation}\label{sec3.1}

Based on the clinical plan evaluation criteria summarized in Table~\ref{tab:clinical plan evaluation template}, we formulate the proposed CDM loss that directly optimizes the specified dose metrics. It incorporates both \textit{D--} and \textit{V--metrics}, allowing end-to-end training guided by clinical evaluation criteria. The following sections describe the differentiable formulation of \textit{D--metrics} and the surrogate-based differentiable approximation of \textit{V--metrics}.

\begin{figure}[tb]
    \centering
    \includegraphics[scale=0.45]{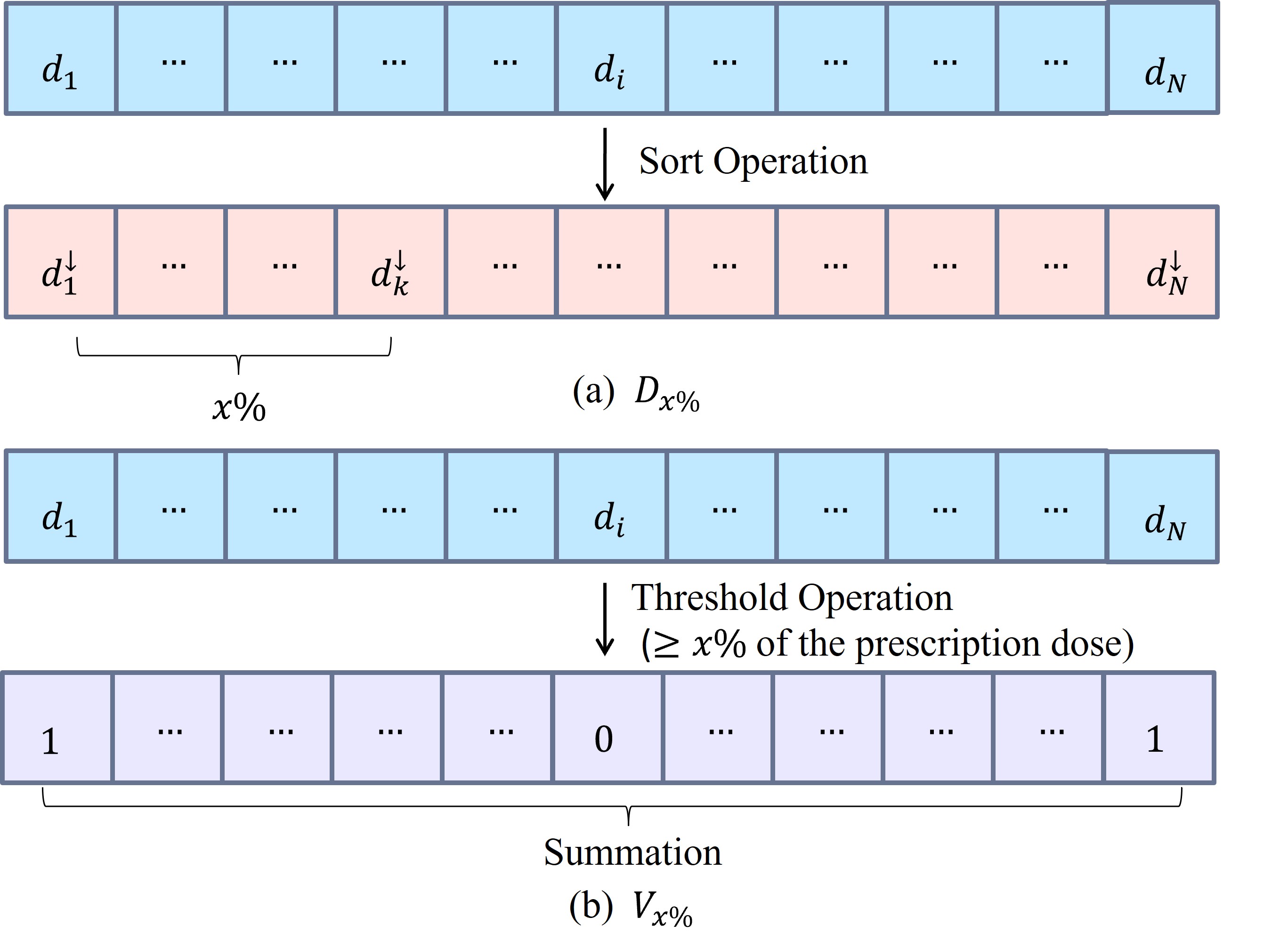}
    \caption{(a) The 3D dose distribution within a ROI is flattened into a one-dimensional dose array $\{d_i\}_{i=1}^{N}$. After a descending sorting operation, we get $d^{\downarrow}_{1} \ge d^{\downarrow}_{2} \ge \dots \ge d^{\downarrow}_{N}$. A \textit{D--metric}, such as $D_{x\%}$, is obtained by retrieving the dose value $d^{\downarrow}_k$ at 
    the index $k$ that corresponds to the top $x\%$ portion of the ROI volume in the sorted dose array. 
   (b) Using the same flattened dose array $\{d_i\}_{i=1}^{N}$, a \textit{V--metric} such as $V_{x\%}$ is computed by applying a dose threshold equal to $x\%$ of the prescription dose, summing the number of voxels that exceed this threshold, and 
    computing the corresponding volume percentage. 
    For example, $V_{95\%}$ of $\mathrm{PTV}_{70}$ denotes the fraction of the $\mathrm{PTV}_{70}$ volume 
    receiving at least $95\%$ of the 70~Gy dose level.
}
    \label{fig:DV}
\end{figure}

\subsubsection{Differentiable Formulation of \textit{D--metrics}}

As shown in Fig.~\ref{fig:DV}(a), given a set of voxel doses $\{d_i\}_{i=1}^{N}$ within an ROI of size $N$, sorted in descending order as $d^{\downarrow}_{1} \ge d^{\downarrow}_{2} \ge \dots \ge d^{\downarrow}_{N}$, the minimum dose received by the hottest $x\%$ of the ROI volume, is denoted as: 
\begin{equation}
    D_{x\%} = d^{\downarrow}_{k}, 
     ~~\text{where } k = \left\lceil \frac{x}{100}N \right\rceil.
\end{equation}
Thus, $D_{x\%}$ corresponds to the $k$-th largest dose value, which can be found at index $k$ after sorting. Similarly, the minimum dose received by the hottest $x$~cc of the ROI, denoted as $D_{x~\text{cc}}$, is computed in an analogous manner:
\begin{equation}
    D_{x~\text{cc}} = d^{\downarrow}_{k}, 
    ~~\text{where } k = \left\lceil \frac{x}{V_\text{voxel}} \right\rceil,
\end{equation}
where $V_\text{voxel}$ denotes the volume of a single voxel (in~cc). Both $D_{x\%}$ and $D_{x~\text{cc}}$ can therefore be interpreted as quantile-based metrics, differing only in whether the cutoff is defined by a fractional or absolute volume.

In addition to these quantile metrics, maximum and minimum dose within a ROI can be expressed naturally in the same framework:
\begin{equation}
D_{\text{max}} = d^{\downarrow}_{1},~~
D_{\text{min}} = d^{\downarrow}_{N}.
\end{equation}

Although sorting indices are non-differentiable,  deep learning frameworks such as \texttt{PyTorch}\footnote{\url{https://pytorch.org/}} allow gradients to propagate through the sorted values via gather/scatter operations.
Following prior work~\cite{wang2022deep, teng2024beam}, we use a top-$k$ algorithm that computes the required quantiles without explicitly sorting all voxels, offering both differentiability and computational efficiency.

Finally, the mean dose $D_{\mathrm{mean}} = \frac{1}{N} \sum_{i=1}^{N} d_i$ is inherently differentiable.

\subsubsection{Differentiable Surrogate of \textit{V--metrics}}

\textit{V--metrics}, including $V_{x\%}$ and $V_{x~\text{Gy}}$, describe the fraction of an ROI receiving at least a specified dose level. As illustrated in Fig.~\ref{fig:DV}(b), $V_{x\%}$ denotes the fraction of the ROI receiving at least $x\%$ of the prescription dose. Given the prescription dose $D_{\text{presc}}$, the corresponding threshold is $T = x D_{\text{presc}} / 100$ Gy. The computation of $V_{x\%}$ is equivalent to counting the number of voxels with dose $d_i \geq T$, normalized by the total number of voxels $N$:
\begin{equation}
    V_{x\%} = \frac{1}{N} \sum_{i=1}^{N} H(d_i - T),
\end{equation}
where $H(\cdot)$ is the non-differentiable Heaviside step function. The same principle applies to $V_{x~\text{Gy}}$, differing only in that the threshold is specified in absolute dose.

To enable gradient-based training, we approximate $H$ with a logistic sigmoid $\sigma_{\alpha}(t) = 1/(1+e^{-\alpha t})$, with slope parameter $\alpha > 0$:
\begin{equation}
    V_{x\%}^{\mathrm{approx}} = \frac{1}{N} \sum_{i=1}^{N} 
    \sigma_{\alpha}\left(d_i - T\right).
\end{equation}

A larger $\alpha$ causes $\sigma_{\alpha}$ to more closely approximate the hard threshold, but could also increase the risk of gradient saturation: when $\alpha$ is too large, most voxels have $|\alpha(d_i - T)| \gg 1$, placing them in the flat tails of the sigmoid where $\sigma'_\alpha(t) \approx 0$. Consequently, only voxels very close to $T$ contribute meaningful gradients, which may destabilize training or slow convergence. Therefore, rather than arbitrarily increasing $\alpha$, we seek the smallest value that ensures sufficient approximation accuracy while mitigating the risk of gradient saturation.

\subsubsection{Selection of $\alpha$ via a Bound on the Approximation Error}

Let $z_i = d_i - T$.  We first quantify how closely the smooth surrogate
$\sigma_\alpha(z)$ matches the hard indicator $H(z)$ at the voxel level.  The mean pointwise approximation error is defined as:
\begin{equation}
    \delta(\alpha) = \frac{1}{N} \sum_{i=1}^N |H(z_i) - \sigma_\alpha(z_i)|.
\end{equation}
Since
\begin{equation}
    V_{x\%} = \frac{1}{N} \sum_{i=1}^N H(z_i),
    \qquad
    V_{x\%}^{\mathrm{approx}}(\alpha)
    = \frac{1}{N} \sum_{i=1}^N \sigma_\alpha(z_i),
\end{equation}
the triangle inequality implies that the $V_{x\%}$ approximation error is always bounded by
\begin{equation}
    \label{eq:V-error-leq-delta}
    \big| V_{x\%} - V_{x\%}^{\mathrm{approx}}(\alpha) \big|
    = \left| \frac{1}{N}\sum_{i=1}^N (H(z_i) - \sigma_\alpha(z_i)) \right|
    \le \delta(\alpha).
\end{equation}
We now derive an upper bound for $\delta(\alpha)$.  
Using the identity
\begin{equation}
    |H(z) - \sigma_\alpha(z)| =
    \begin{cases}
        1 - \sigma_\alpha(z) = \dfrac{1}{1+e^{\alpha z}}, & z \ge 0, \\[6pt]
        \sigma_\alpha(z)   = \dfrac{1}{1+e^{-\alpha z}}, & z < 0,
    \end{cases}
\end{equation}
we obtain the symmetric form
\begin{equation}
    |H(z) - \sigma_\alpha(z)|
    = \frac{1}{1 + e^{\alpha |z|}}
    \le e^{-\alpha |z|}.
\end{equation}
A voxel-wise application yields
\begin{equation}
    \delta(\alpha)
    \le \frac{1}{N} \sum_{i=1}^N e^{-\alpha |z_i|}.
\end{equation}

To obtain an interpretable and clinically meaningful bound, let $q_m$ denote the fraction of voxels whose dose values lie within a margin $m$ around the threshold:
\begin{equation}
    q_m = \frac{1}{N} \#\{\, i : |z_i| \le m \,\}.
\end{equation}
For voxels inside this margin, the worst-case error is $1/2$ (since
$\sigma_\alpha(0)=1/2$).  
For voxels outside the margin, the error is bounded by $e^{-\alpha m}$.  
Thus both $\delta(\alpha)$ and the approximation error of $V_{x\%}$  in Eq.
\eqref{eq:V-error-leq-delta} satisfy
\begin{equation}
    \big| V_{x\%} - V_{x\%}^{\mathrm{approx}}(\alpha) \big|
    \le \delta(\alpha)
    \le q_m\!\left(\tfrac12\right)
    + (1 - q_m) e^{-\alpha m}.
\end{equation}

To ensure that the total approximation error remains below a user-specified
tolerance $\varepsilon$ (\%), it suffices that
\begin{equation}
    q_m\!\left(\tfrac12\right)
    + (1 - q_m)e^{-\alpha m}
    \le \varepsilon.
\end{equation}
Solving for $\alpha$ gives the required lower bound
\begin{equation}
    \alpha \ge \frac{1}{m}
    \ln\!\left(
        \frac{1 - q_m}{\varepsilon - \tfrac{1}{2}q_m}
    \right).
    \label{eq:alpha_min}
\end{equation}

Eq.~\eqref{eq:alpha_min} provides a lower bound on the slope parameter, ensuring a balance
between approximation accuracy and avoidance of gradient saturation.

In summary, for each \textit{V--metric} we first choose a tolerance $\varepsilon$
that serves as an upper bound on the error tolerated by the user.
We then define a margin $m$ around the dose threshold and compute $q_m$,
the fraction of voxels whose dose values fall within this margin,
from the dose distributions of the training set.
Finally, we take the smallest $\alpha$ satisfying Eq.~\eqref{eq:alpha_min} and use it in the loss calculation.

\subsubsection{Overall Loss Formulation}

Using the differentiable formulation of \textit{D--metrics} and the differentiable surrogate of \textit{V--metrics} described above, we compute each metric in the template (Table~\ref{tab:clinical plan evaluation template}) for both the predicted and ground truth dose in the same way. The loss is then defined as:
\begin{equation}
    \mathcal{L}_{\mathrm{CDM}} =
    \sum_{r \in \mathcal{R}}
    \sum_{k \in \mathcal{M}_r}
    w_{r,k}\,
    \big|
    M^{\mathrm{pred}}_{r,k}
    -
    M^{\mathrm{gt}}_{r,k}
    \big|,
\end{equation}
where $M^{\mathrm{pred}}_{r,k}$ and $M^{\mathrm{gt}}_{r,k}$ denote the predicted and ground truth metric $k$ for ROI $r$, and $w_{r,k}$ a weight. To preserve voxel-wise accuracy of the predicted dose distribution, we also use the MAE loss:
\begin{equation}
    \mathcal{L}_{\mathrm{MAE}} =
    \frac{1}{|\Omega|}
    \sum_{v \in \Omega}
    \big|
    D^{\mathrm{pred}}(v)
    -
    D^{\mathrm{gt}}(v)
    \big|.
\end{equation}

The final objective combines both components:
\begin{equation}
    \mathcal{L}_{\mathrm{total}} =
    \lambda_1 \mathcal{L}_{\mathrm{MAE}}
    +
    \lambda_2 \mathcal{L}_{\mathrm{CDM}},
\end{equation}
where $\lambda_1$ and $\lambda_2$ balance voxel-wise consistency and alignment of clinical objectives.

\begin{figure}[tb]
    \centering
    \includegraphics[scale=0.45]{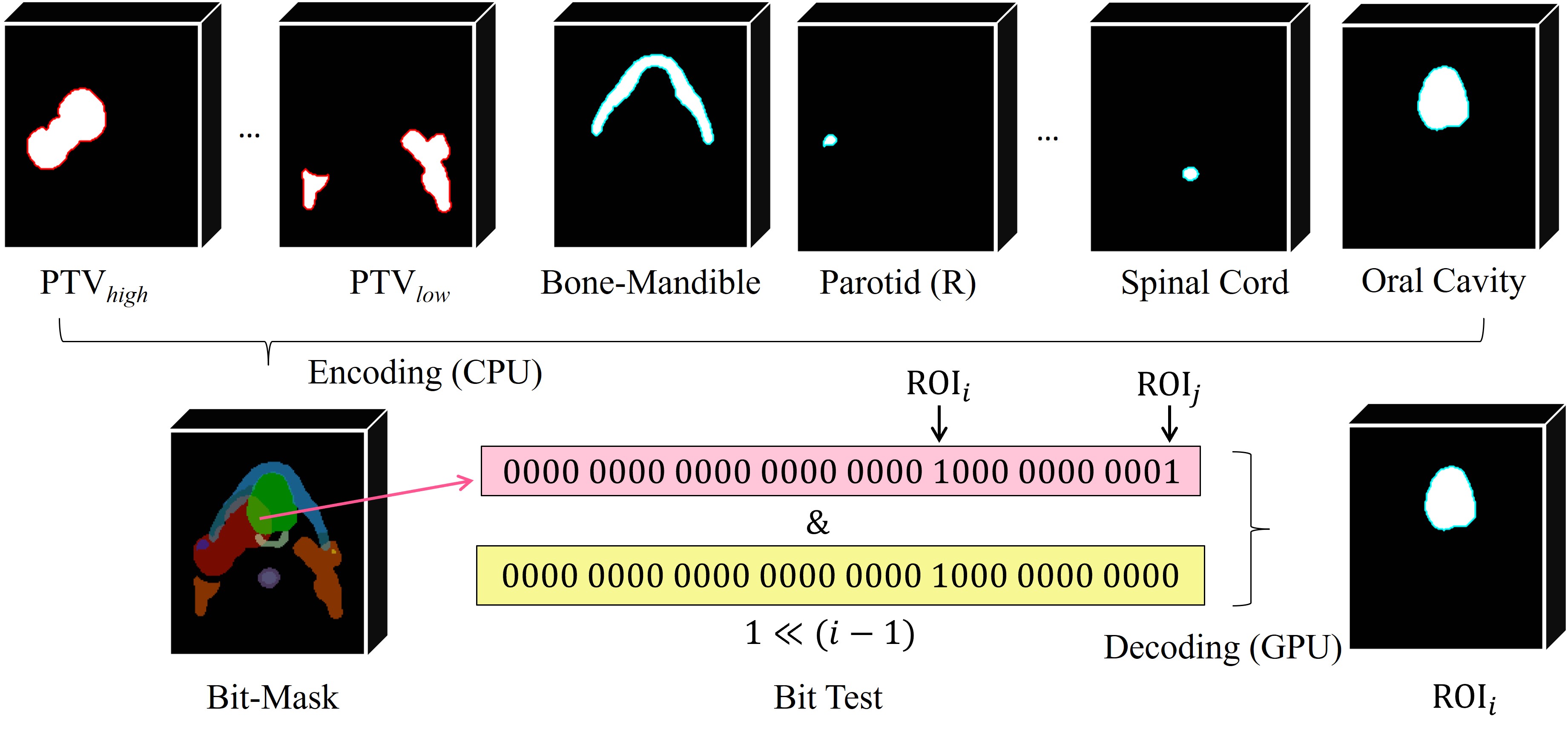}
    \caption{
    Multiple individual ROI masks are losslessly encoded into a single integer bit-mask, where each bit position represents a specific ROI (illustrated in pink). Because the raw bit-mask values span a wide numerical range, direct visualization is not intuitive. Therefore, the bit-mask is visualized in pseudo-color, with colors assigned solely based on their unique values for display purposes. During decoding, the binary mask for ROI$_i$ is obtained by testing whether the $(i\!-\!1)$-th bit of the bit-mask is active, implemented as a bitwise AND operation on GPU between the bit-mask and $(1 \ll (i-1))$, where the left-shifted single-bit mask $(1 \ll (i-1))$ is illustrated in yellow. A nonzero result indicates voxel membership in ROI$_i$.
}
    \label{fig:Bitmask}
\end{figure}

\subsection{Bit-Mask Encoding}\label{sec3.2}

The CDM loss in Section~\ref{sec3.1} requires access to all ROIs
listed in Table~\ref{tab:clinical plan evaluation template}. In H\&N RT, the number of ROIs is typically large (often exceeding twenty), and many of them overlap spatially. In conventional dose prediction pipelines, each ROI is stored as an independent 3D binary mask channel using a one--hot encoding. When full 3D volumes are used as network input, this representation leads to substantial overhead in CPU preprocessing and CPU–-to–-GPU data transfer.

To alleviate these inefficiencies, we introduce a lossless bit–mask encoding as a compact intermediate representation for ROI handling, as illustrated in Fig.~\ref{fig:Bitmask}. Importantly, the bit-mask is not used as the direct input to the network.
Instead, it serves as an efficient representation where standard
binary ROI masks are decoded when needed, either to form the network input or to
compute ROI-specific loss terms.

Formally, let $S_i(\mathbf r)\in\{0,1\}$ denote voxel membership in ROI$_i$.
Assigning one bit position to each ROI yields the lossless encoding
\begin{equation}
    B(\mathbf{r}) =
    \sum_{i=1}^{N_{\mathrm{ROI}}}
        S_i(\mathbf{r})\, 2^{\,i-1},
    \qquad N_{\mathrm{ROI}} \le B_{\mathrm{max}},
\end{equation}
where $B_{\mathrm{max}}$ denotes the bit width of the integer type.
In this work, a 32-bit unsigned integer (\texttt{uint32}) is used, supporting up to 32 ROIs. Overlapping anatomical structures are represented by activating multiple bits within the same voxel, preserving all ROI relationships in a single channel. 

A key advantage of this representation arises during data preprocessing. With a conventional one--hot encoding, geometric augmentations such as rotations, flips, and affine transformations must be applied independently to each ROI channel, causing preprocessing cost to scale linearly with the number of ROIs.
In contrast, when ROIs are stored in a single bit-mask volume, each augmentation operation is applied once to the bit-mask, implicitly transforming all ROIs simultaneously. This substantially reduces CPU preprocessing time while maintaining exact ROI geometry.

The bit-mask representation also reduces CPU–-to–-GPU data transfer. Instead of transferring the CT volume together with more than twenty one--hot ROI channels, only the CT volume and a single compact bit-mask are transferred per iteration. This reduces I/O overhead and improves training throughput, especially when using full-resolution 3D volumes.

Once on the GPU, the bit–mask $B$ is decoded into binary ROI masks as required. When forming the network input, the decoded ROI masks 
$\hat{S}\in \mathbb{R}^{N_{\mathrm{ROI}}\times D\times H\times W}$ are concatenated with the CT volume to obtain
\begin{equation}
    I_{\mathrm{in}}
    \in
    \mathbb{R}^{(1+N_{\mathrm{ROI}})\times D\times H\times W}.
\end{equation}
Thus, the network always operates on standard one--hot ROI channels, identical to those used in conventional pipelines. The bit-mask itself serves only as an intermediate encoding and does not alter the inputs of the network.

The decoding is implemented via lightweight, element-wise bit-test, as shown in Fig.~\ref{fig:Bitmask}. In particular, the
binary mask for ROI$_i$ is obtained via
\begin{equation}
    S_i(\mathbf r)
    = \mathrm{bool}\!\left(
        B(\mathbf r)\;\&\;(1 \ll (i-1))
    \right),
\end{equation}
where \(\mathrm{bool}(\cdot)\) maps nonzero values to~1, ``\&'' denotes the bitwise AND, and ``$\ll$'' the bit left-shift operation. These operations are highly efficient on modern
GPUs and introduce negligible computational overhead.

During loss computation, only the bit–mask $B$ remains resident in GPU memory, and the binary mask for each
ROI$_i$ is decoded on demand at the moment its corresponding metric
$M^{\mathrm{pred}}_{r,k}$ or $M^{\mathrm{gt}}_{r,k}$ is evaluated. The decoded mask is used to gather voxel doses for that ROI and is then immediately
released.
This on--demand decoding strategy avoids storing all $N_{\mathrm{ROI}}$ ROI channels
simultaneously and thereby reduces peak GPU memory usage.

\section{Experiment and Results}

\subsection{Experiment Setup}\label{sec:Experiment Setup}

The ROIs used in our experiments are listed in Table~\ref{tab:clinical plan evaluation template}.
The template contains two PTV structures ($\mathrm{PTV}_{54.25}$ and $\mathrm{PTV}_{70}$) and
28 OARs, resulting in a total of $N_{\mathrm{ROI}} = 30$ structures.
Because this number fits within a 32-bit range, all ROI masks are encoded using a 
32-bit unsigned integer (\texttt{uint32}) following the bit–mask representation described in 
Section~\ref{sec3.2}.

For the CDM loss, all PTV-related $V_{95\%}$ metric were included in the differentiable surrogate of \textit{V--metrics}.
To ensure controlled approximation accuracy, we adopt a margin of $m=0.5$\,Gy (i.e., a 1 Gy window) and a tolerance of $\varepsilon=1\%$. Using the lower-bound condition in Eq.~(\ref{eq:alpha_min}), we compute
the minimum $\alpha$ values that satisfy the prescribed tolerance.
Thus, we set $\alpha = 209$ for $\text{PTV}_{54.25}$ and 
$\alpha = 176$ for $\text{PTV}_{70}$. To reflect the higher clinical priority of PTVs, all PTV-related metric weights are set to $w(r,k)=1$, whereas all OAR-related metric weights are set to $w(r,k)=0.1$. The balancing coefficients are set to 
$\lambda_1 = 1$ and $\lambda_2 = 0.5$ for the final loss
$\mathcal{L}_{\mathrm{total}}$.

Unless otherwise stated, all experiments were conducted using a 3D U-Net baseline implemented in the 
\texttt{MONAI}\footnote{\url{https://github.com/Project-MONAI/MONAI}} framework. 
The architecture follows the standard encoder–decoder design with skip connections~\cite{falk2019u}. 
Compared with the conventional 3D U-Net, the major architectural modification is the use of  3D Instance Normalization instead of Batch Normalization, which has been shown to provide
more stable training for small-batch 3D medical imaging DL tasks~\cite{kolarik2020comparing}.

All models were trained for 1000 epochs using the AdamW optimizer with an initial learning rate of $3\times10^{-4}$ and a cosine annealing schedule. Mixed-precision training (bf16-mixed) was used throughout to reduce memory consumption and improve computational efficiency. All experiments were performed on an NVIDIA RTX~6000 Ada GPU.

\subsection{Results}

\begin{table}[tb]
\caption{Comparison of model performance under different loss-function configurations. Lower scores indicate better performance.}
\label{tab:loss_exp}
\centering
\renewcommand{\arraystretch}{1.1}
\begin{tabular}{llll}
\toprule
Loss function & PTV Score~(\%)~$\downarrow$  & OAR Score~(Gy)~$\downarrow$ & Dose Score~(Gy)~$\downarrow$\\
\midrule
MAE & 1.544 $\pm$ 1.188 & 2.103 $\pm$ 0.704 & \textbf{1.300 $\pm$ 0.229} \\
MAE + DVH & 0.992 $\pm$ 0.466 & 2.162 $\pm$ 0.518 & 1.385 $\pm$ 0.217 \\ 
MAE + DVH + CDM & 0.567 $\pm$ 0.300 & \textbf{1.933 $\pm$ 0.481} & 1.389 $\pm$ 0.228 \\
MAE + CDM (proposed) & \textbf{0.491 $\pm$ 0.250} & 1.999 $\pm$ 0.497 & 1.370 $\pm$ 0.245 \\
\bottomrule
\end{tabular}
\end{table}

\begin{figure}[tb]
    \centering
    \includegraphics[scale=0.55]{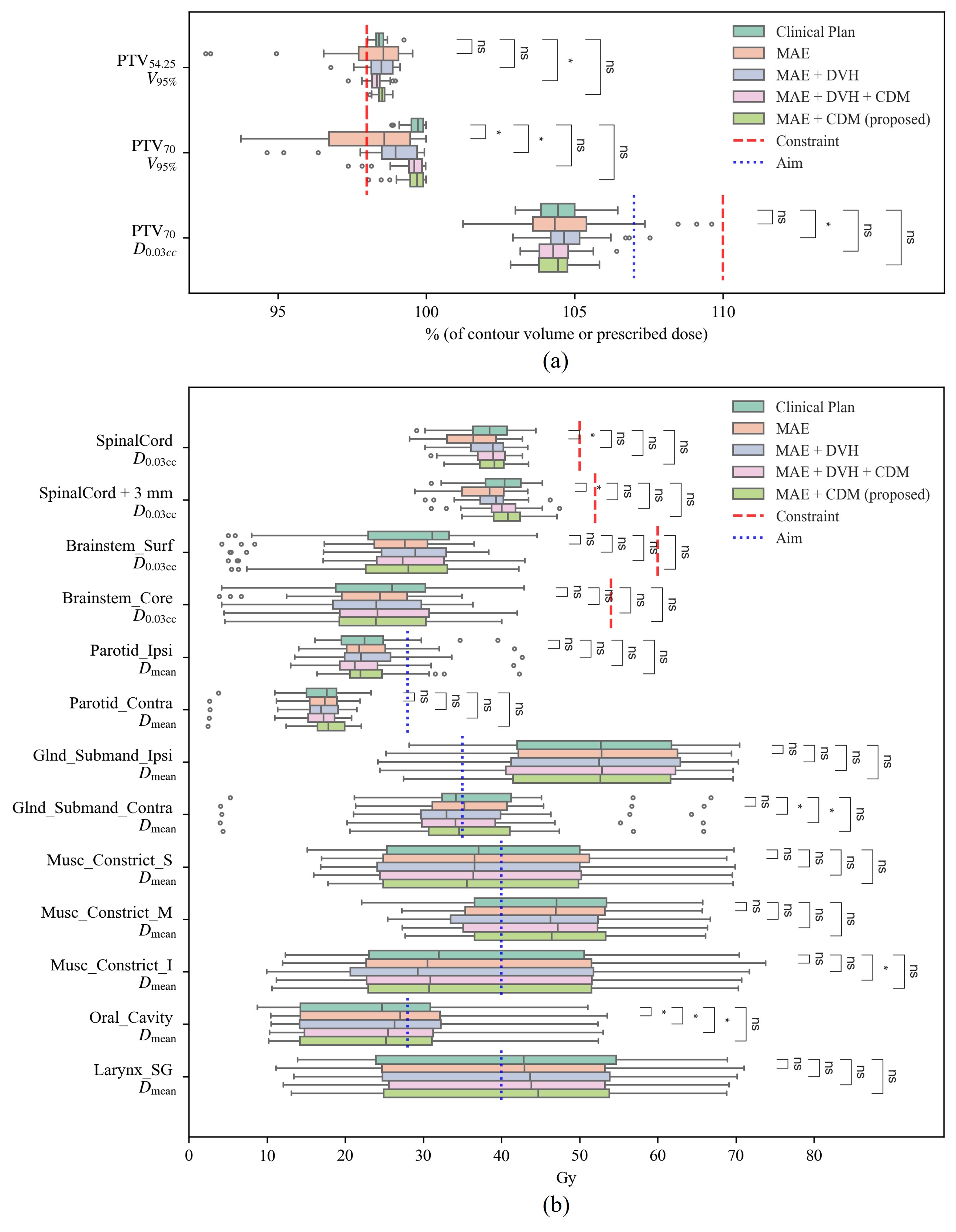}
    \caption{Boxplots of dose evaluation metrics for the clinically most relevant ROIs for different loss function configurations. (a) Boxplots for PTVs. (b) Boxplots for OARs. The corresponding planning aims and dose constraints are also shown (aims are clinically regularly violated due to proximity of the tumor). Parotid and submandibular glands were categorized as ipsilateral (Ipsi) or contralateral (Contra) according to received dose. Statistical significance was tested using a two-tailed Wilcoxon signed-rank test. *: $p \leq$ 0.05, ns = not significant.}
    \label{fig:constraint_boxplot}
\end{figure}

\subsubsection{Loss Function Comparison}
We first compared different loss-function configurations using the 3D U-Net baseline to evaluate the impact of the proposed CDM loss.  Four loss-function configurations were investigated: MAE, MAE + DVH, MAE + DVH + CDM, and MAE + CDM (our proposed setting). Here, the DVH loss refers to the DVH value loss introduced in~\cite{wang2022deep}, which provides a more efficient way to penalize DVH-curve discrepancies compared with the formulation in~\cite{nguyen2020incorporating}. In all experiments, the MAE loss was weighted as 1, while both the DVH and CDM components were assigned weights of 0.5. Table~\ref{tab:loss_exp} reports the model performance under different loss-function configurations. The results show that introducing the CDM loss led to a substantial improvement in the PTV Score, achieving the lowest value when combined with MAE. In contrast, both MAE and MAE + DVH alone failed to effectively reduce the PTV Score. The CDM loss also provided moderate benefits for the OAR Score, suggesting its positive influence on balancing target coverage and organ-at-risk sparing. Although the MAE-only configuration yielded the lowest Dose Score, it also had very high PTV Score.

\begin{figure}[tb]
    \centering
    \includegraphics[scale=0.40]{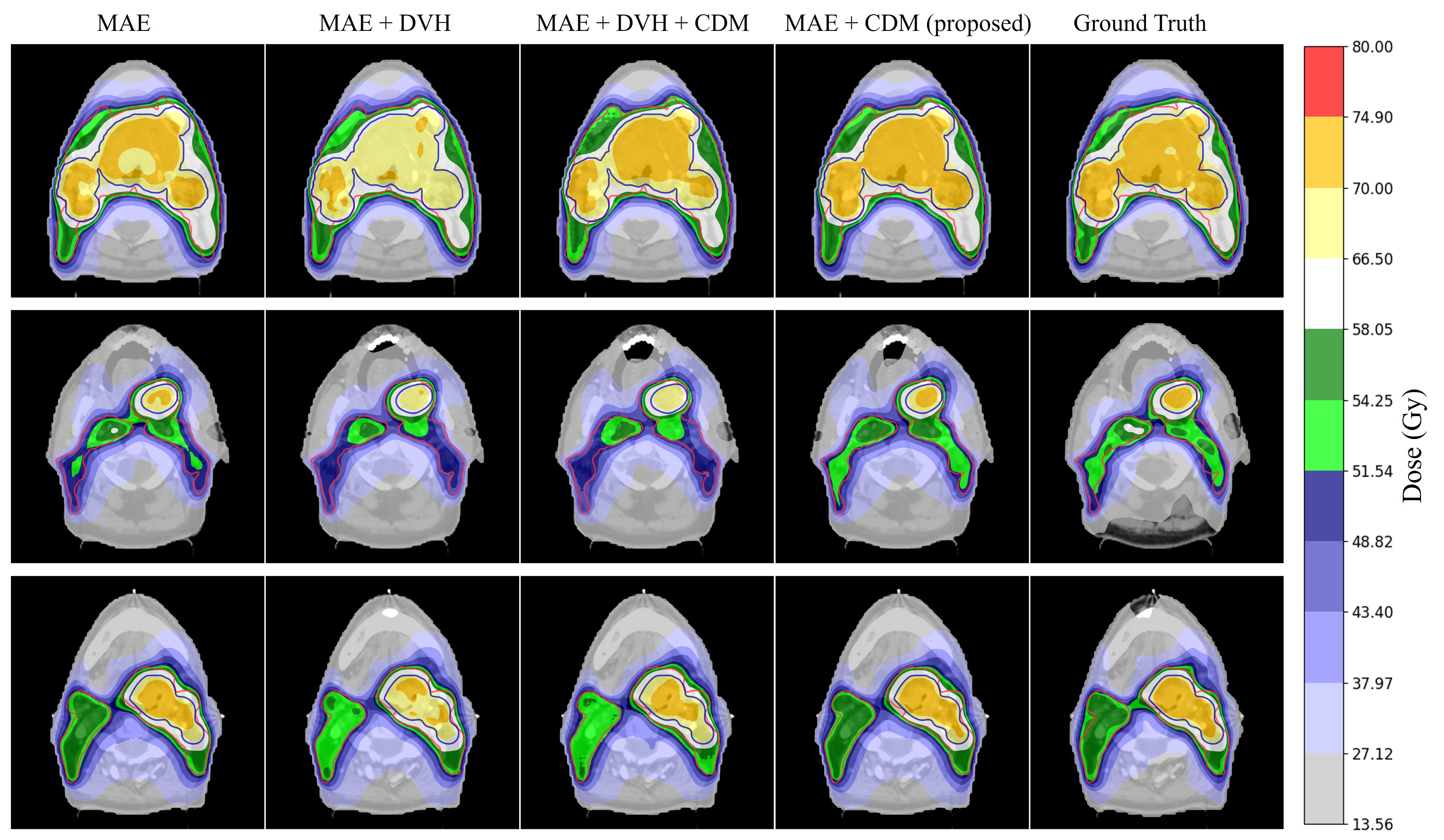}
    \caption{Axial visualization of dose distributions under different loss function configurations. The red contour denotes the $\text{PTV}_{54.25}$, while the blue contour represents the $\text{PTV}_{70}$. A colormap routinely used in clinical practice is employed to enhance visual interpretability.}
    \label{fig:dose_compare_visualization}
\end{figure}

We further examined whether the predicted dose distributions satisfied the predefined planning constraints. As illustrated in Fig.~\ref{fig:constraint_boxplot}, the MAE + CDM configuration was the only setting that met both the $V_{95\%}$ ($\geq 98\%$) and $D_{0.03 cc}$ ($\leq 110\%$) constraints across all test cases. No statistically significant differences from the clinical plans were observed for these metrics. In contrast, the other loss configurations showed instances of insufficient PTV coverage or excessive dose, as shown in Fig.~\ref{fig:dose_compare_visualization}. For OARs, all loss configurations satisfied the relevant clinical limits, but only the MAE + CDM model showed no statistically significant differences from the clinical plans for all metrics.

\subsubsection{Ablation Study}

To assess the influence of $\alpha$ used in the differentiable \textit{V--metric} approximation, we performed an ablation study on the parameter $\alpha$ (Table~\ref{tab:alpha_ablation}). Smaller values (e.g., 100) resulted in higher PTV Scores. The theoretically derived values $(209, 176)$ for $\text{PTV}_{54.25}$ and $\text{PTV}_{70}$ achieved the best PTV Score while maintaining stable OAR and Dose Scores. Increasing $\alpha$ beyond this range (300–500) did not yield further improvement and in some cases slightly degraded PTV performance.

To further assess the computational impact of the proposed bit–mask ROI encoding, an ablation study was performed using the same 3D U-Net baseline and identical training settings. The comparison is reported in Table~\ref{tab:bitmask_ablation}. When using conventional one-hot ROI channels, training required a peak GPU memory of 20.50~GB and an average epoch duration of 241~s. With bit–mask encoding, the peak memory was 19.57~GB (4.5\% reduction) and the epoch duration was 43~s (82.2\% reduction). Peak GPU memory is reported because memory usage during training is dynamic, and the maximum allocation determines whether the model fits within available GPU memory.

\begin{table}[tb]
\centering
\caption{Ablation study on the parameter $\alpha$. Lower scores indicate better performance.}
\label{tab:alpha_ablation}
\renewcommand{\arraystretch}{1.1}
\begin{tabular}{ccccc}
\toprule
$\alpha~(\text{PTV}_{54.25})$ & $\alpha~(\text{PTV}_{70})$ & PTV Score~(\%)~$\downarrow$  & OAR Score~(Gy)~$\downarrow$ & Dose Score~(Gy)~$\downarrow$\\
\midrule
100 & 100 & 0.564 $\pm$ 0.287 & \textbf{1.919 $\pm$ 0.502} & 1.349 $\pm$ 0.228 \\
209 & 176  & \textbf{0.491 $\pm$ 0.250} & 1.999 $\pm$ 0.497 & 1.370 $\pm$ 0.245 \\
300 & 300 & 0.500 $\pm$ 0.276 & 1.961 $\pm$ 0.491 & \textbf{1.346 $\pm$ 0.226} \\
400 & 400 & 0.563 $\pm$ 0.298 & 1.988 $\pm$ 0.475 & 1.352 $\pm$ 0.214 \\
500 & 500 & 0.563 $\pm$ 0.308 & 1.993 $\pm$ 0.452 & 1.327 $\pm$ 0.204 \\
\bottomrule
\end{tabular}
\end{table}

\begin{table}[tb]
\centering
\caption{Ablation experiment on the 3D U-Net baseline comparing computational efficiency with and without the bit–mask ROI encoding.}
\label{tab:bitmask_ablation}
\renewcommand{\arraystretch}{1.1}
\begin{tabular}{lcc}
\toprule
Method (3D U-Net baseline) & Peak GPU Memory (GB) & Epoch Duration (s) \\
\midrule
Without bit–mask & 20.50 & 241 \\
With bit–mask (proposed) & \textbf{19.57} & \textbf{43} \\
\bottomrule
\end{tabular}
\end{table}

\subsubsection{Model Comparison}

\begin{table}[tb]
\caption{Comparison of dose prediction performance across different network architectures using the MAE + CDM loss configuration. Lower scores indicate better performance.}
\label{tab:model_exp}
\centering
\renewcommand{\arraystretch}{1.1}
\begin{tabular}{lccc}
\toprule
Model & PTV Score~(\%)~$\downarrow$  & OAR Score~(Gy)~$\downarrow$ & Dose Score~(Gy)~$\downarrow$\\
\midrule
SwinUNETR \cite{hatamizadeh2021swin} & 0.699 $\pm$ 0.292 & 2.035 $\pm$ 0.566 & 1.461 $\pm$ 0.253 \\
C3D \cite{liu2021cascade} & 0.497 $\pm$ 0.314 & \textbf{1.945 $\pm$ 0.473} & \textbf{1.289 $\pm$ 0.192} \\
MedNeXt \cite{roy2023mednext}& 0.496 $\pm$ 0.251 & 2.154 $\pm$ 0.571 & 1.395 $\pm$ 0.204 \\
Pyfer \cite{gheshlaghi2024cascade}& \textbf{0.491 $\pm$ 0.287} & 2.019 $\pm$ 0.517 & 1.290 $\pm$ 0.207 \\
3D UNet (baseline) \cite{falk2019u} & \textbf{0.491 $\pm$ 0.250} & 1.999 $\pm$ 0.497 & 1.370 $\pm$ 0.245 \\
\bottomrule
\end{tabular}
\end{table}

Building upon the best-performing loss setting (MAE + CDM), finally we compared the 3D U-Net baseline with several state-of-the-art architectures (Table~\ref{tab:model_exp}), including transformer-based models (SwinUNETR~\cite{hatamizadeh2021swin}), large-kernel ConvNeXt-style convolutional networks (MedNeXt~\cite{roy2023mednext}), and cascade-style frameworks (C3D~\cite{liu2021cascade}, Pyfer~\cite{gheshlaghi2024cascade}). Notably, the 3D U-Net achieved a PTV Score of 0.491, matching the best-performing model (Pyfer) and outperforming more complex architectures such as SwinUNETR and MedNeXt. The OAR Score of 1.999 was also competitive, ranking second among all models.

\section{Discussion}

Overall, our results demonstrate that directly optimizing clinically used dose metrics is substantially more effective than voxel-wise or DVH-curve–based supervision. As shown in Table~\ref{tab:loss_exp}, MAE alone achieved the lowest voxel-wise Dose Score, but failed to ensure adequate target coverage. Previously proposed DVH-based losses improved DVH curve similarity~\cite{nguyen2020incorporating,jhanwar2022domain}. However they did not reliably enforce clinically mandatory constraints, such as $V_{95\%}$ for PTVs. This limitation is evident in Fig.~\ref{fig:constraint_boxplot} and Fig.~\ref{fig:dose_compare_visualization}. Both MAE and MAE + DVH exhibit cases of insufficient target coverage or excessive hot spots. In contrast, the proposed CDM loss directly optimizes the dose metrics used in clinical plan evaluation. This task-aligned supervision consistently satisfied all PTV constraints in all patients, without compromising OAR sparing. These results indicate that the proposed CDM loss improves clinical relevance and outperforms loss formulations based on voxel-wise or DVH-curve supervision.

A key component of the proposed framework is the differentiable surrogate for \textit{V--metrics}, which uses a sigmoid slope parameter $\alpha$. The analytically derived lower bound provides a principled and reproducible way to select $\alpha$ without requiring exhaustive tuning. The ablation study~(Table~\ref{tab:alpha_ablation}) shows that a small $\alpha$ leads to overly smooth threshold approximations. As a result, the model struggled to distinguish doses above and below the clinical prescription dose level. In contrast, very large $\alpha$ values created overly sharp transitions. This limits the number of voxels that contribute meaningful gradients and makes it harder for the model to correct small deviations near the prescription threshold. Between these extremes, the analytically derived values $(209,176)$ in this study provided stable performance. Additionally, the results also show that slightly larger values such as $\alpha=300$ achieve similar PTV, OAR, and Dose Scores. Accordingly, the chosen values should be interpreted as a conservative and principled choice, rather than an attempt to identify an optimal setting.

An important advantage of the proposed CDM loss is its flexible, template-driven design. All optimized metrics, ROIs, and their relative priorities are specified through a simple JSON-based clinical evaluation template. This allows the loss to be configured for different planning protocols without modifying the network architecture or training pipeline. Although this study focuses on a single-institution H\&N dataset, the template-based formulation in principle can be adapted to other datasets by updating the clinical criteria template. Compared with fixed, hard-coded loss formulations, this approach provides a practical and transparent way to align training objectives with clinically used evaluation standards.

Beyond loss formulation, this work addresses a practical but often overlooked bottleneck in H\&N dose prediction: the computational overhead caused by a large number of ROIs. The proposed bit-mask encoding strategy replaces more than twenty one-hot ROI channels with a single integer-valued mask and decodes ROI membership on the GPU. This substantially reduces training time, with the average epoch duration decreasing from 241~s to 43~s (Table~\ref{tab:bitmask_ablation}), accelerating training by more than $5\times$. In contrast to conventional pipelines
\cite{babier2021openkbp,liu2021cascade,gheshlaghi2024cascade,gao2025factors} that often restrict supervision to a subset of structures due to computational constraints, the proposed encoding enables more comprehensive and clinically faithful supervision. This supports more informative learning of spatial dose relationships across PTVs and OARs.

An additional observation from the model comparison experiments is that architectural complexity becomes a secondary factor when the training objective is well aligned with clinical goals. Under the MAE + CDM loss, a standard 3D U-Net achieved performance comparable to, or better than, transformer-based~\cite{hatamizadeh2021swin} and cascade-style architectures~\cite{liu2021cascade,gheshlaghi2024cascade}~(Table~\ref{tab:model_exp}). This suggests that, for 3D dose prediction, improving \emph{what the model is optimized for} may be more impactful than increasing architectural complexity. From a clinical deployment perspective, this is an encouraging result, as simpler architectures are typically easier to train, validate, and maintain.

Several limitations of this study should be acknowledged. First, all data were obtained from a single institution with a unified planning protocol. The generalizability of the approach to other institutions therefore remains to be validated. Second, while the predicted dose distributions satisfied all clinical constraints, this study did not evaluate downstream integration into deliverable treatment planning~\cite{bakx2021development,borderias2023machine}. Future work will focus on multi-institutional validation and on assessing the impact of CDM-guided dose prediction on dose mimicking and final plan quality.

In summary, this work presents a flexible and clinically aligned framework for 3D head and neck dose prediction that combines a clinical DVH metric loss with an efficient, lossless ROI encoding strategy.
By directly optimizing clinically relevant evaluation metrics, the proposed approach satisfied all predefined clinical constraints and improves the clinical acceptability of the predicted dose distributions.
As such, the framework provides a practical foundation for AI-assisted radiotherapy workflows and may facilitate automated planning by improving plan consistency and reducing manual effort.


\FloatBarrier
\bibliography{sn-bibliography}

@inproceedings{roy2023mednext,
  title={{MedNeXt}: transformer-driven scaling of {ConvNets} for medical image segmentation},
  author={Roy, Saikat and Koehler, Gregor and Ulrich, Constantin and Baumgartner, Michael and Petersen, Jens and Isensee, Fabian and Jaeger, Paul F and Maier-Hein, Klaus H},
  booktitle={International Conference on Medical Image Computing and Computer-Assisted Intervention},
  pages={405--415},
  year={2023},
  organization={Springer}
}

@inproceedings{hatamizadeh2021swin,
  title={{Swin UNETR}: swin transformers for semantic segmentation of brain tumors in {MRI} images},
  author={Hatamizadeh, Ali and Nath, Vishwesh and Tang, Yucheng and Yang, Dong and Roth, Holger R and Xu, Daguang},
  booktitle={International MICCAI Brainlesion Workshop},
  pages={272--284},
  year={2021},
  organization={Springer}
}

@article{berry2016interobserver,
  title={Interobserver variability in radiation therapy plan output: results of a single-institution study},
  author={Berry, Sean L and Boczkowski, Amanda and Ma, Rongtao and Mechalakos, James and Hunt, Margie},
  journal={Practical Radiation Oncology},
  volume={6},
  number={6},
  pages={442--449},
  year={2016},
  publisher={Elsevier}
}

@article{falk2019u,
  title={{U-Net}: deep learning for cell counting, detection, and morphometry},
  author={Falk, Thorsten and Mai, Dominic and Bensch, Robert and {\c{C}}i{\c{c}}ek, {\"O}zg{\"u}n and Abdulkadir, Ahmed and Marrakchi, Yassine and B{\"o}hm, Anton and Deubner, Jan and J{\"a}ckel, Zoe and Seiwald, Katharina and others},
  journal={Nature Methods},
  volume={16},
  number={1},
  pages={67--70},
  year={2019},
  publisher={Nature Publishing Group US New York}
}

@article{bakx2021development,
  title={Development and evaluation of radiotherapy deep learning dose prediction models for breast cancer},
  author={Bakx, Nienke and Bluemink, Hanneke and Hagelaar, Els and van der Sangen, Maurice and Theuws, Jacqueline and Hurkmans, Coen},
  journal={Physics and Imaging in Radiation Oncology},
  volume={17},
  pages={65--70},
  year={2021},
  publisher={Elsevier}
}

@article{buciuman2022adaptive,
  title={Adaptive radiotherapy in head and neck cancer using volumetric modulated arc therapy},
  author={Buciuman, Nikolett and Marcu, Loredana G},
  journal={Journal of Personalized Medicine},
  volume={12},
  number={5},
  pages={668},
  year={2022},
  publisher={MDPI}
}

@article{mcintosh2015contextual,
  title={Contextual atlas regression forests: multiple-atlas-based automated dose prediction in radiation therapy},
  author={McIntosh, Chris and Purdie, Thomas G},
  journal={IEEE Transactions on Medical Imaging},
  volume={35},
  number={4},
  pages={1000--1012},
  year={2015},
  publisher={IEEE}
}

@article{liu2021cascade,
  title={A cascade 3D {U-Net} for dose prediction in radiotherapy},
  author={Liu, Shuolin and Zhang, Jingjing and Li, Teng and Yan, Hui and Liu, Jianfei},
  journal={Medical Physics},
  volume={48},
  number={9},
  pages={5574--5582},
  year={2021},
  publisher={Wiley Online Library}
}

@article{gheshlaghi2024cascade,
  title={A cascade transformer-based model for 3D dose distribution prediction in head and neck cancer radiotherapy},
  author={Gheshlaghi, Tara and Nabavi, Shahabedin and Shirzadikia, Samireh and Moghaddam, Mohsen Ebrahimi and Rostampour, Nima},
  journal={Physics in Medicine \& Biology},
  volume={69},
  number={4},
  pages={045010},
  year={2024},
  publisher={IOP Publishing}
}

@article{lin2024lenas,
  title={{LENAS}: Learning-Based Neural Architecture Search and Ensemble for {3D} Radiotherapy Dose Prediction},
  author={Lin, Yi and Liu, Yanfei and Chen, Hao and Yang, Xin and Ma, Kai and Zheng, Yefeng and Cheng, Kwang-Ting},
  journal={IEEE Transactions on Cybernetics},
  year={2024},
  publisher={IEEE}
}

@article{gao2025factors,
  title={On factors that influence deep learning-based dose prediction of head and neck tumors},
  author={Gao, Ruochen and Mody, Prerak and Rao, Chinmay and Dankers, Frank and Staring, Marius},
  journal={Physics in Medicine \& Biology},
  volume={70},
  number={11},
  pages={115006},
  year={2025},
  publisher={IOP Publishing}
}

@inproceedings{wang2022deep,
  title={Deep learning-based head and neck radiotherapy planning dose prediction via beam-wise dose decomposition},
  author={Wang, Bin and Teng, Lin and Mei, Lanzhuju and Cui, Zhiming and Xu, Xuanang and Feng, Qianjin and Shen, Dinggang},
  booktitle={International Conference on Medical Image Computing and Computer-Assisted Intervention},
  pages={575--584},
  year={2022},
  organization={Springer}
}

@article{teng2024beam,
  title={Beam-wise dose composition learning for head and neck cancer dose prediction in radiotherapy},
  author={Teng, Lin and Wang, Bin and Xu, Xuanang and Zhang, Jiadong and Mei, Lanzhuju and Feng, Qianjin and Shen, Dinggang},
  journal={Medical Image Analysis},
  volume={92},
  pages={103045},
  year={2024},
  publisher={Elsevier}
}

@article{nguyen2020incorporating,
  title={Incorporating human and learned domain knowledge into training deep neural networks: a differentiable dose-volume histogram and adversarial inspired framework for generating Pareto optimal dose distributions in radiation therapy},
  author={Nguyen, Dan and McBeth, Rafe and Sadeghnejad Barkousaraie, Azar and Bohara, Gyanendra and Shen, Chenyang and Jia, Xun and Jiang, Steve},
  journal={Medical Physics},
  volume={47},
  number={3},
  pages={837--849},
  year={2020},
  publisher={Wiley Online Library}
}

@article{jhanwar2022domain,
  title={Domain knowledge driven {3D} dose prediction using moment-based loss function},
  author={Jhanwar, Gourav and Dahiya, Navdeep and Ghahremani, Parmida and Zarepisheh, Masoud and Nadeem, Saad},
  journal={Physics in Medicine \& Biology},
  volume={67},
  number={18},
  pages={185017},
  year={2022},
  publisher={IOP Publishing}
}

@article{hancock2020survey,
  title={Survey on categorical data for neural networks},
  author={Hancock, John T and Khoshgoftaar, Taghi M},
  journal={Journal of Big Data},
  volume={7},
  number={1},
  pages={28},
  year={2020},
  publisher={Springer}
}

@article{babier2021openkbp,
  title={{OpenKBP}: the open-access knowledge-based planning grand challenge and dataset},
  author={Babier, Aaron and Zhang, Binghao and Mahmood, Rafid and Moore, Kevin L and Purdie, Thomas G and McNiven, Andrea L and Chan, Timothy CY},
  journal={Medical Physics},
  volume={48},
  number={9},
  pages={5549--5561},
  year={2021},
  publisher={Wiley Online Library}
}

@inproceedings{kolarik2020comparing,
  title={Comparing normalization methods for limited batch size segmentation neural networks},
  author={Kolarik, Martin and Burget, Radim and Riha, Kamil},
  booktitle={2020 43rd international conference on telecommunications and signal processing (TSP)},
  pages={677--680},
  year={2020},
  organization={IEEE}
}

@article{borderias2023machine,
  title={Machine learning-based automatic proton therapy planning: impact of post-processing and dose-mimicking in plan robustness},
  author={Borderias-Villarroel, Elena and Huet Dastarac, Margerie and Barrag{\'a}n-Montero, Ana Mar{\'\i}a and Helander, Rasmus and Holmstrom, Mats and Geets, Xavier and Sterpin, Edmond},
  journal={Medical Physics},
  volume={50},
  number={7},
  pages={4480--4490},
  year={2023},
  publisher={Wiley Online Library}
}

\end{document}